\documentclass[]{bytedance_seed}

\usepackage[toc,page,header]{appendix}

\usepackage{natbib}

\usepackage{CJKutf8}

\usepackage{xargs}  

\usepackage{todonotes}  

\usepackage{multirow}

\usepackage{cleveref}

\usepackage{amsmath}
\usepackage{dsfont}

\usepackage{subcaption}

\usepackage{svg}

\usepackage{mathrsfs}
\usepackage{adjustbox}
\usepackage{multirow}
\usepackage{multirow}
\usepackage{multicol}
\usepackage{tcolorbox}
\usepackage{changepage}
\usepackage{graphicx}
\usepackage{amssymb}
\usepackage{array}
\usepackage{bm}
\usepackage{hyperref}
\usepackage{wrapfig}
\usepackage{bbding}
\usepackage{xurl}

\usepackage{minitoc}

\title{DreaMontage: Arbitrary Frame-Guided \\One-Shot Video Generation}

\author[*\dagger]{\normalsize{Jiawei Liu}}
\author[*]{\normalsize{Junqiao Li}}
\author[*\dagger]{\normalsize{Jiangfan Deng}}
\author[*]{\normalsize{Gen Li}}
\author[]{\normalsize{Siyu Zhou}}
\author[]{\normalsize{Zetao Fang}}
\author[]{\normalsize{Shanshan Lao, \\ Zengde Deng}}
\author[]{\normalsize{Jianing Zhu}}
\author[]{\normalsize{Tingting Ma}}
\author[]{\normalsize{Jiayi Li}}
\author[]{\normalsize{Yunqiu Wang}}
\author[]{\normalsize{Qian He}}
\author[]{\normalsize{Xinglong Wu}}
\affiliation[]{\normalsize{Intelligence Creation Team, ByteDance}}
\abstract{
The "one-shot" technique represents a distinct and sophisticated aesthetic in filmmaking. 
However, its practical realization is often hindered by prohibitive costs and complex real-world constraints. 
Although emerging video generation models offer a virtual alternative, existing approaches typically rely on naïve clip concatenation, which frequently fails to maintain visual smoothness and temporal coherence. 
In this paper, we introduce DreaMontage, a comprehensive framework designed for arbitrary frame-guided generation, capable of synthesizing seamless, expressive, and long-duration one-shot videos from diverse user-provided inputs. 
To achieve this, we address the challenge through three primary dimensions. 
(i) We integrate a lightweight intermediate-conditioning mechanism into the DiT architecture.
By employing an Adaptive Tuning strategy that effectively leverages base training data, we unlock robust arbitrary-frame control capabilities. 
(ii) To enhance visual fidelity and cinematic expressiveness, we curate a high-quality dataset and implement a Visual Expression SFT stage.
In addressing critical issues such as subject motion rationality and transition smoothness, we apply a Tailored DPO scheme, which significantly improves the success rate and usability of the generated content. 
(iii) To facilitate the production of extended sequences, we design a Segment-wise Auto-Regressive (SAR) inference strategy that operates in a memory-efficient manner. 
Extensive experiments demonstrate that our approach achieves visually striking and seamlessly coherent one-shot effects while maintaining computational efficiency, empowering users to transform fragmented visual materials into vivid, cohesive one-shot cinematic experiences.
}

\checkdata[Date]{December 25, 2025}
\checkdata[Project Page]{\url{ https://dreamontage.github.io/DreaMontage} }

\begin{document}
\begin{CJK*}{UTF8}{gbsn}

\maketitle
% =========================================================
% 【页脚代码】必须紧跟在 \maketitle 后面
% =========================================================
{
  % 临时将脚注符号改为特殊的符号序列 (*, †, ‡, §, ...)
  \renewcommand{\thefootnote}{\fnsymbol{footnote}}
  % [1] 代表第一个符号 (*)，对应 Equal contributions
  \footnotetext[1]{\footnotesize Equal contributions}
  % [2] 代表第二个符号 (†)，对应通讯作者和邮箱
  \footnotetext[2]{\footnotesize Corresponding authors: \texttt{\{liujiawei.cc22, dengjiangfan\}@bytedance.com}}
}
% =========================================================

\definecolor{chinese_red}{HTML}{8B4513}
\definecolor{english_blue}{HTML}{4169E1}

% \begin{figure}[ph]
% \begin{center}
% % \vspace{-30pt}
% \includegraphics[height=8.0cm]{figures/teaser.png}
% % \includegraphics[trim=0cm 23cm 0cm 0cm,clip,height=6.5cm]{figures/radar_new/radar_aa_arena.png}
% \end{center}
% \label{fig:teaser}
% \caption{\textbf{DreaMontage}. DreaMontage could generate long one-shot video (up-to 60s) guided by arbitrary images or videos at any timestep.}
% \end{figure}%

% \clearpage

% \tableofcontents

% \newpage

\section{Introduction}
\begin{figure}[t]
\begin{center}
% \vspace{-30pt}
\includegraphics[width=\textwidth]{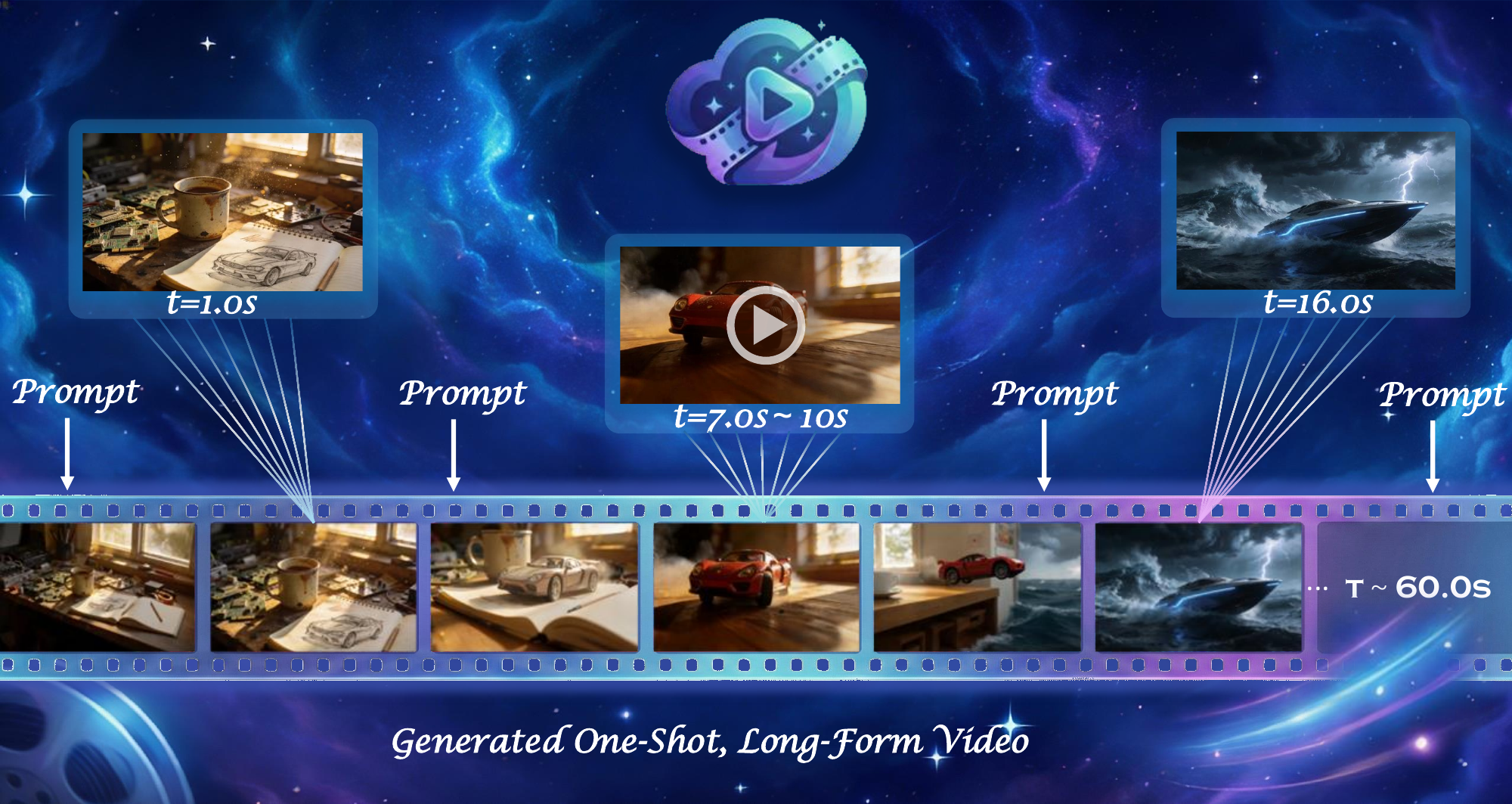}
\end{center}
\label{fig:teaser}
\caption{\textbf{DreaMontage: Flexible Dreams, Seamless Montage.} Our model generates one-shot, long-form videos guided by arbitrary keyframes or video clips anchored at precise temporal locations.}
\end{figure}%

\begin{figure*}[t]
\includegraphics[width=\textwidth]{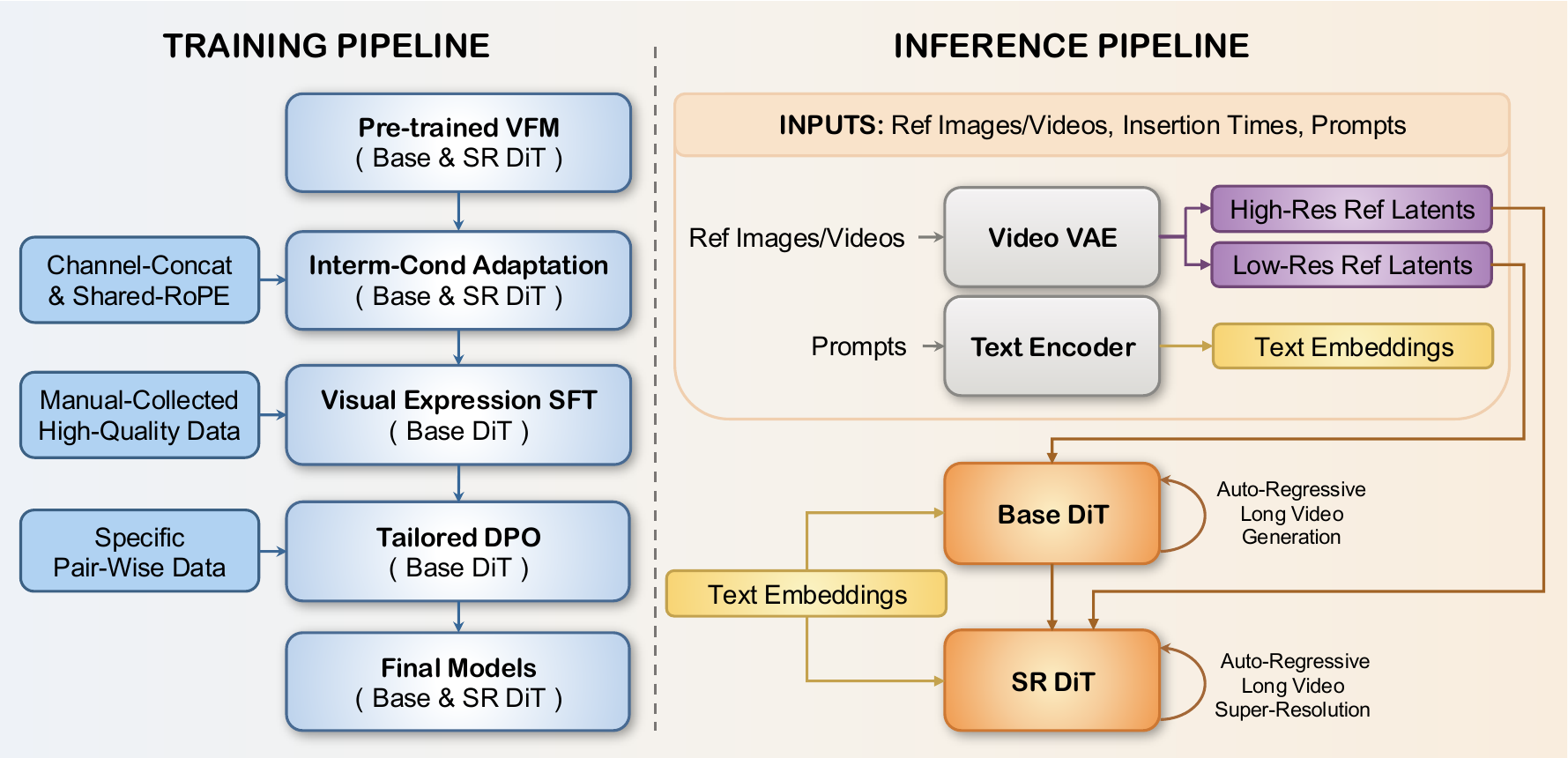}
\caption{\textbf{Overview of the DreaMontage.} 
          The left panel illustrates the multi-stage training pipeline, progressing from the Adaptive Tuning to the Visual Expression SFT and Tailored DPO. The right panel depicts the inference pipeline, where reference (condition) images/videos and rephrased prompts guide the generation process, supporting auto-regressive long-video generation.}
\label{fig:overview}
\end{figure*}
The "one-shot" (or "long take") technique represents a distinct aesthetic orientation in filmmaking, celebrated for its immersive continuity. However, executing a physical one-shot video incurs substantial costs in set design, staging, and post-production, while demanding exceptional professional skill. Furthermore, traditional filmmaking is strictly bound by physical space limitations, constraining imaginative scope and creative freedom.
Recently, the burgeoning capabilities of video generation models~\cite{openai2024sora, google2024veo,kuaishou2024kling,gao2025seedance,wan2025wan,kong2024hunyuanvideo,peng2025open} have opened new avenues for this task, potentially allowing users to leverage existing visual materials to produce coherent one-shot long videos. A prevalent practice involves concatenating multiple clips generated via first-last frame conditioning. However, this approach fails to fundamentally guarantee the smoothness and coherence of the video content, often resulting in disjointed transitions. Consequently, an arbitrary frame-conditioning video generator has become a critical requirement. Ideally, such a system should produce seamless, cohesive, and engaging one-shot videos while affording precise temporal control.

We first formulate an explicit definition of this task: given a set of images and/or video clips alongside their temporal positions, the model generates a single continuous shot that obeys user instructions and ensures coherent transitions between conditioning contents. 
Typically, modern video generation models comprise a Variational Autoencoder (VAE)~\cite{kingma2013auto} and a Diffusion Transformer (DiT)~\cite{peebles2023scalable}. 
From a technical perspective, equipping existing frameworks with such capabilities faces three primary obstacles. 
First, the 3D VAE~\cite{openai2024sora} encoder usually adopts a causal mechanism with temporal down-sampling. 
Consequently, intermediate reference images cannot be intuitively represented in the latent space, hindering precise frame-level control. 
Second, the content between conditioning frames often exhibits significant semantic or visual shifts, posing severe challenges to the base model's continuity and leading to undesirable abrupt cuts. 
Third, one-shot videos inherently require longer durations, whereas modern DiT-based models demand substantial memory and computational resources.
Satisfying these overheads for long-video generation is non-trivial.

In this work, we propose DreaMontage, an arbitrary frame-guided one-shot video generator. 
To address the aforementioned challenges, we introduce innovations across three dimensions.
As demonstrated in Fig.~\ref{fig:overview},
first, an intermediate-conditioning mechanism is designed: we follow the channel-wise concatenation mode used in the typical image-to-video task and propose an extra Shared-RoPE condition strategy in the super-resolution DiT.
By filtering the base model's training data, we employ an Adaptive Training scheme to mitigate frame-wise misalignment, thereby equipping the generator with basic arbitrary frame-conditioning capabilities.
Second, to enhance the expressiveness of one-shot generation, we carefully categorize video types and curate a small-scale, high-quality dataset. 
We then leverage Supervised Fine-Tuning (SFT) for visual expression.
To tackle issues like abrupt cuts and unnatural subject motion, we construct specific pairwise datasets and apply Direct Preference Optimization (DPO)~\cite{rafailov2023direct,xiao2024cal,wallace2024diffusion}.
Finally, for inference, we design a Segment-wise Auto-Regressive (SAR) generation mechanism. This strategy decouples long video generation from strict computational and memory constraints while preserving the integrity of the one-shot content.
Through this holistic design, DreaMontage empowers creators to orchestrate complex narratives where disparate visual assets merge into a unified sequence, maintaining high visual fidelity and temporal coherence.

In summary, our contributions are mainly three-fold: 
\begin{itemize}
 \item We propose a simple yet efficient method for inserting intermediate conditions—including independent frames and video clips, unlocking arbitrary frame-conditioning capabilities within a DiT-based model at a relatively low cost.
 \item We curate high-quality datasets and implement a progressive training pipeline involving SFT and DPO, empowering the model to generate seamless, coherent, and vivid one-shot videos.
 \item We design a specialized Segment-wise Auto-Regressive (SAR) generation mechanism to enable long one-shot video production, striking an optimal balance between performance and efficiency.
\end{itemize}

% \iffalse

\section{Related Work}
With the rapid advancement of diffusion models, the performance of AI-driven video generation has been continuously evolving. Commercial systems such as Sora~\cite{openai2024sora}, Veo~\cite{google2024veo}, Kling~\cite{kuaishou2024kling}, Seaweed~\cite{seawead2025seaweed7bcosteffectivetrainingvideo} and Seedance~\cite{gao2025seedance}, as well as open-source counterparts like Wan~\cite{wan2025wan}, Hunyuan Video~\cite{kong2024hunyuanvideo}, and CogVideoX~\cite{yang2024cogvideox}, have demonstrated remarkable video quality and temporal consistency. 
Among various downstream tasks~\cite{guo2023i2v,girdhar2023emu,MakePixelsDance,chen2023seine,chen2025humo,liu2025phantom,feng2024i2vcontrol,feng2025i2vcontrol}, image-to-video (I2V) generation has attracted particular attention due to its controllability. 
However, relying solely on the first and/or last frame as the conditioning signal remains restrictive, lacking fine-grained control over motion evolution and intermediate transformations. 
To address this limitation, we extend the task boundaries, enabling the model to support combinations of multiple image and video conditions that can be placed at arbitrary positions along the temporal axis. 
This enhances the foundation model’s flexibility and controllability, allowing it to generate long take videos with richer visual variations.

In the I2V task, different foundation models adopt distinct mechanisms for injecting conditioning. 
Models such as Open-Sora~\cite{zheng2024open}, LTX-Video~\cite{hacohen2024ltx} and LongCat~\cite{team2025longcat} employ timestep-based conditioning control, where the noise tokens and conditioning tokens are assigned different timesteps, enabling the model to learn to distinguish conditioning signals based on timestep information. 
Beyond timestep-based conditioning, Hunyuan Video further introduces in-context control for image conditions. It integrates a semantic image injection module to extract the semantic tokens from the input conditioning image, which are then injected through a combination of dual-stream and single-stream DiT blocks. 
This approach increases the length of the contextual token sequence, which not only raises computational costs but also limits the feasibility of using long sequences as conditioning inputs. 
Models such as Wan and Open-Sora Plan~\cite{lin2024open} prepare guidance frames that match the shape of the target video, with non-conditioning regions set to zero, and then feed them into the VideoVAE to obtain the latent representation. 
Consequently, regardless of how many conditioning frames are used, the model must encode a conditional video of the same length as the target video. 
Due to the use of Conv3D in the VideoVAE, this results in high computational cost and substantial redundant computation. 
In contrast, we adopt a more efficient approach by extracting the VAE latent from the condition images/videos and concatenating it with the noise latent along the channel dimension. 
This approach is simple and intuitive, and our experiments demonstrate that this channel-wise concatenation mechanism can effectively handle the multi-intermediate conditions.
\begin{figure}[t]
\centering
\includegraphics[width=\textwidth]{figures/base.pdf}
\caption{\textbf{The Interm-Cond Adaptation strategy.} 
            (a) Due to the Causalty VAE's temporal downsampling, an intermediate latent aggregates information from multiple frames, making it an imprecise condition for a specific timestamp. 
            (b) To resolve this, we align the training distribution with inference. 
            Each single condition frame (or the initial frame of a condition video) is re-encoded while the subsequent frames of the condition video are re-sampled from the latent distribution.}
\label{fig:base}
\end{figure}

\section{Method}
\label{sec:method}

\subsection{Interm-Cond Adaptation}
In this work, we adopt Seedance 1.0~\cite{gao2025seedance}, a video generation framework based on DiT architecture. As illustrated in the right half of Fig.~\ref{fig:overview}, the pipeline starts from a 3D Video Variational Autoencoder (VideoVAE) compressing images and videos into a compact latent space. A text encoder is utilized to encode textual information, which is subsequently integrated into the DiT backbone via cross-attention. The generation process consists of two steps: first, a base DiT model generates video latents at a low resolution (480p); then, a super-resolution (SR) DiT model enhances the primary latents to the desired higher resolution (720p/1080p). In DreaMontage, our modifications are aimed at facilitating intermediate conditioning (Interm-Cond) and ensuring the efficacy of arbitrary-frame guided generation.

\subsubsection{Model Design}
\noindent\textbf{Base Model.}
In the original image-to-video task of the base model, the initial frame is channel-wise concatenated on the noise latent. We follow the same approach when conditioning intermediate frames. However, due to the causality inherent in the VAE encoder, this channel-wise concatenation mode encounters an correspondence issue. As illustrated in Fig.~\ref{fig:base}-(a), in case of the independent frame conditioning, its VAE latent corresponds to multiple frames of the video to be generated. In the case of video conditioning (which can be deemed as multiple continuous frames) , the first frame faces a similar dilemma. In our work, we find that this problem can largely be solved through lightweight tuning with the Interm-Cond setup. Therefore, we construct a one-shot subset from the base model's training data, and then devise an efficient Adaptive Tuning strategy to enable the generation from intermediately inserted images and videos (details provided in the next section).

\begin{wrapfigure}{r}{0.5\textwidth}
    \centering
    \vspace{10pt} % 调整图片上方的空白（可选，视情况调整）
    \includegraphics[width=\linewidth]{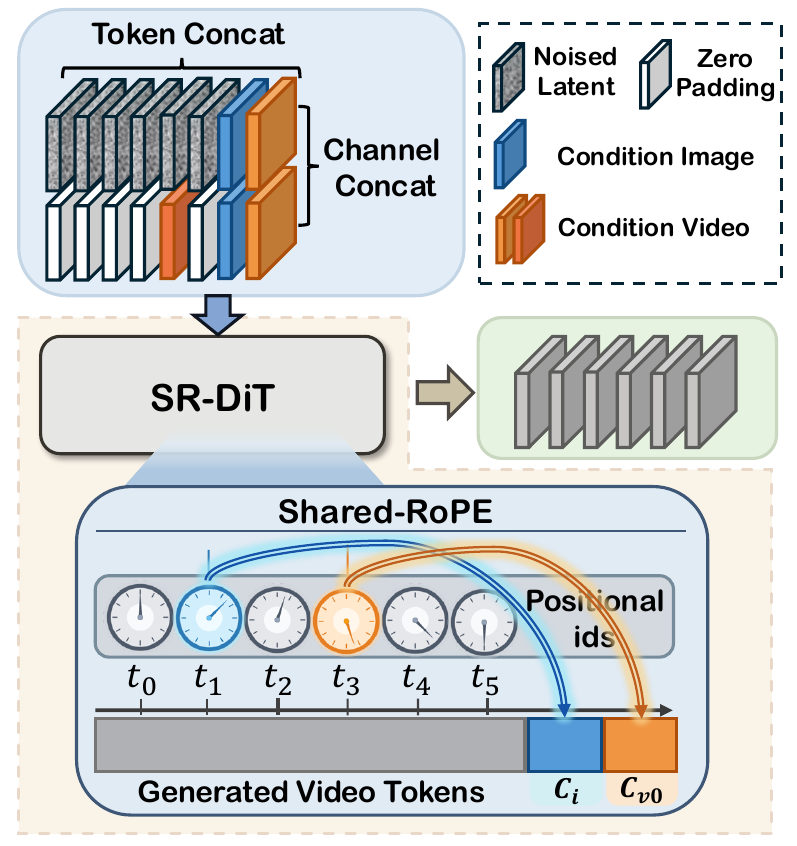}
    \caption{\textbf{The Shared-RoPE strategy for the super-resolution model.} 
             In addition to channel-wise concatenation, we introduce a sequence-wise conditioning mechanism to eliminate artifacts. 
             Condition frames are appended to the tail of the sequence while share the same RoPE value as the target frames they guide (e.g., $C_i$ shares the RoPE of $t_1$). 
             In the case of video condition, this strategy is only applied to the first frame.
         }
    \label{fig:sr}
    \vspace{-30pt} % 调整caption下方的空白
\end{wrapfigure}

\noindent\textbf{Super-Resolution Model.}
 In the super-resolution (SR), conditional frames are inserted as signals for high-resolution guidance. However, in case of Interm-Cond, we frequently observe flickering and cross-frame color shifts. These artifacts are primarily attributed to the amplification of discrepancies between the "conditions" and the "generated contents" by the SR model, in which the conditions are channel-wise concatenated. In order to mitigate this problem, we propose Shared-RoPE: for each reference image, besides the channel-wise conditioning, an extra sequence-wise conditioning is applied. As shown in Fig.~\ref{fig:sr}, the VAE latent is directly concatenated along the noise sequence, with the value of Rotary Position Embedding (RoPE)~\cite{su2024roformer} being set the same as those at the corresponding position. Especially, in the case of video condition, the Shared-RoPE is only applied to the first frame to avoid excessive computational overhead.
 % {r}表示图片靠右 (right)，{l}则靠左
% {0.5\textwidth} 表示图片占用当前文本宽度的50%，也可以设为具体数值如 {6cm}

\subsubsection{Adaptive Tuning}
\noindent\textbf{Data Filtering.}
We start by constructing a one-shot subset from the base corpus through a data filtering pipeline. Specifically, a VLM-based scene detection model is adopted to exclude multi-shot videos. The cosine similarity is calcuated between the CLIP~\cite{radford2021learning} features of the first and last frame for each video, filtering out those with high similarity to retain videos exhibiting large visual variations. We use Q-Align~\cite{wu2023q} to assess aesthetic scores, thereby eliminating data with low aesthetic quality. To ensure adequate motion intensity, a 2D optical flow predictor is applied to estimate the video’s motion strength. Additionally, the RTMPose~\cite{jiang2023rtmpose} is utilized to filter out high-quality human-centric videos with clear pose structure. Through this meticulous data filtering process, we finally obtain a one-shot subset featuring large variations, strong motion, and high aesthetic quality. To emphasize the multi-action characters in the one-shot data above, we adopt an internally trained action recognizer to identify action intervals within each video, followed by a VLM captioner that generates dense descriptions for each action, resulting in structured, action-wise video annotations.

\noindent\textbf{Training.}
During training, we propose an efficient way to support arbitrary conditions and make full use of the data. 
Primarily, the VAE latents of all videos are pre-calculated to ensure computational efficiency.
As shown in Fig.~\ref{fig:base}-(b), for single image condition, the boundary frames of each action are treated as conditioning inputs. 
To enhance the input diversity and improve model robustness, we additionally sample random intermediate frames within each video to serve as supplementary conditions.
These frames are re-encoded with the single-image mode by the VAE encoder.
For video condition, segments with varying lengths are randomly extracted from the latents. 
It is noteworthy that these "latent segments" are not a typical encoding result due to the causality of the VAE (the pre-calculated latents are acquired by causal encoding from the start of the whole video). 
We mitigate this problem by adopting an approximate approach: for each segment, the first frame is re-encoded while the subsequent frames are re-sampled from the latent distribution. 

\subsection{Visual Expression SFT}

To further enhance the model’s instruction-following ability and motion expressiveness, we perform Supervised Fine-Tuning on newly collected, category-balanced data with stronger one-shot characteristics. 

\noindent\textbf{Data Collection.}
We begin by conducting a fine-grained analysis of the model’s underperforming cases and categorizing them into five major classes: Camera Shots, Visual Effects, Sport, Spatial Perception, and Advanced Transitions. Each major class is further divided into multiple subclasses to ensure precise targeting. For instance, the "Camera Shots" class includes subclasses such as "Basic Camera Movements – Dolly In" and "Shooting Technique – First-Person View (FPV)", while the "Visual Effects" class covers "Generation – Light" and "Transformation – Animal Metamorphosis". This hierarchical classification system sets a clear direction for subsequent data collection. Based on the taxonomy, we collect a small-scale dataset consisting of videos for each subclass. Samples in this dataset are carefully selected to capture the core characteristics of each scenario while prioritizing high motion dynamics. Compared with the adaptive tuning data, videos for SFT have longer durations (up to 20 seconds) and cover a greater number of seamless scene transitions.

\noindent\textbf{Training.} 
The SFT is conducted upon the model weights from the previous adaptive tuning. We follow the similar training strategies and random condition settings in the adaptive tuning stage. After the SFT, the model acquires remarkable enhancement for motion dynamics and instruction-following of the video contents.

\subsection{Tailored DPO}
\begin{figure*}[t]
\centering
\includegraphics[width=\textwidth]{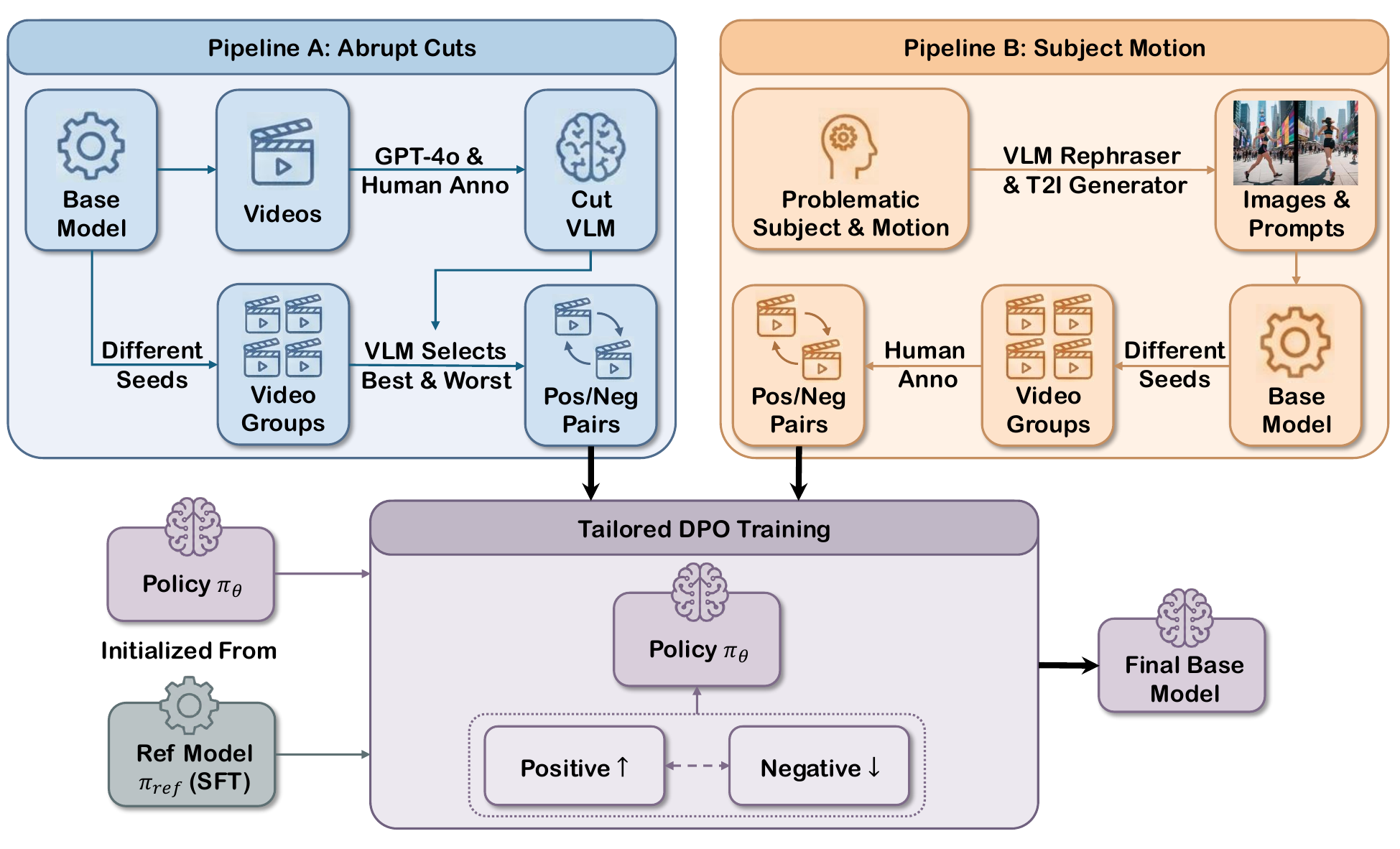}
\caption{\textbf{Illustration of the Tailored DPO.} 
To eliminate specific generation artifacts, we construct preference pairs via two distinct pipelines: \textbf{Pipeline A} addresses abrupt cuts by leveraging a trained VLM discriminator to automatically select positive/negative samples, whereas \textbf{Pipeline B} targets subject motion rationality through human-annotated screening of challenging cases. These pairs subsequently drive the DPO training to optimize the policy $\pi_\theta$ against the reference model $\pi_{\text{ref}}$, ensuring smoother transitions and physically plausible motions.}
\label{fig:dpo}
\end{figure*}

The model encounters two typical issues when generating one-shot videos. On the one hand, if there is a significant difference between adjacent conditions, the video generated may experience undesired abrupt cuts. On the other hand, motion of subjects (such as human, animals and vehicles) in the video often deviates from physical laws and is prone to distortions. To mitigate these problems, we design tailored pipelines to produce contrastive pair-data and conduct specified Direct Preference Optimization (DPO)~\cite{rafailov2023direct,wallace2024diffusion} training, as shown in Fig.~\ref{fig:dpo}.

\noindent\textbf{Abrupt Cuts.}
The basic idea of solving the "abrupt cuts" issue is guiding the model to learn a preference against cuts. We make two steps in achieving this goal. First, a vision-language model (VLM) is trained to recognize abrupts cuts within a video. Specifically, we generate a dataset comprising 10k video clips using the base model through the fist-last frame conditioning (the simplest special case of arbitray-conditioning), where each clip is categorized into five distinct levels based on "the severity of the cuts". In this process, the initial categorization is performed by GPT-4o~\cite{openai2024gpt4o} and subsequently refined by human annotators. This dataset is then utilized to fine-tune a VLM, enabling the model to effectively identify cuts. After acquisition of this "abrupt cut discriminator", we leverage it to construct pairs of "cut" and "non-cut" videos. Particularly, the base generation model is employed again to produce a large number of video "groups", each containing videos generated from the same prompt (first/last condition frames and texts) but with different seeds. The VLM is then applied to these groups to select the "best" and "worst" videos in terms of the cut severity mentioned above, thereby generating a substantial number of contrastive pairs. Finally, these pairs are utilized in DPO training, steering the generator to avoid producing videos that contain abrupt cuts.

\noindent\textbf{Subject Motion.}
The approach to mitigating the subject motion problem follows a similar path. Primarily, we carefully identify and collect a set of common subjects (humans, vehicles, etc.) along with their frequently occurring problematic actions (jumping, rotating, turning, etc.) within the context of video generation. After that, a text-to-image generator Seedream~\cite{seedream2025seedream} and a VLM-based rephraser are adopted to create a set of image-to-video prompts, each comprising a "first frame", a "last frame" and a "descriptive text". This prompt set is then utilized to generate video groups using the base generation model, adhering to the analogous methodology described above. In contrast to the abrupt cuts issue, we observe that current top vision-language models struggle to recognize the unreasonable motion problem. Therefore, we directly select high-contrast pairs with the assistance of human annotators. The resulting pairs are then incorporated into DPO training, guiding the generator to produce more reasonable motion for the subjects.

\noindent\textbf{Optimization Objective.}
Our Tailored DPO aims to strictly penalize the identified artifacts. 
We skip complex reward modeling and directly optimize the policy $\pi_\theta$ against the reference model $\pi_{\text{ref}}$ (initialized from the SFT weights). 
The objective is to minimize the standard DPO loss:
\begin{equation}
\mathcal{L}_{\text{DPO}} = -\mathbb{E}_{(c, v_w, v_l) \sim \mathcal{D}} \left[ \log \sigma \left( \beta \log \frac{\pi_\theta(v_w|c)}{\pi_{\text{ref}}(v_w|c)} - \beta \log \frac{\pi_\theta(v_l|c)}{\pi_{\text{ref}}(v_l|c)} \right) \right],
\label{eq:dpo}
\end{equation}

where $c$ denotes the visual and textual conditions, and $\beta$ controls the deviation from $\pi_{\text{ref}}$. 
By maximizing the margin between the likelihood of plausible samples ($v_w$) and artifact-prone ones ($v_l$), the model is effectively calibrated to produce coherent long-take videos while preserving the diversity learned during SFT.

\subsection{Segment-wise Auto-Regressive Generation}
To address the challenge of making long one-shot videos that exceed the computational capacity of single-pass generation, we propose Segment-wise Auto-Regressive (SAR) generation strategy. 
Specifically, a variable-length sliding window is applied in the latent space to partition the target video into multiple consecutive segments,
during which the user-provided conditions are treated as candidate boundaries.
As the window slides, the current segment is terminated on the latest boundary once the window exceeds a predetermined maximum length.
After the partitioning, an auto-regressive mechanism is used in successively generating these segments.

Formally, we define the generation of the $n$-th segment $\mathbf{s}_n$ as a conditional mapping process parameterized by $\theta$. Specifically, the model synthesizes the current latent sequence by conditioning on the tail latents of the preceding segment — extracted via a temporal operator $\tau(\cdot)$ — alongside the set of local guidance signals $\mathcal{C}_n$:
\begin{equation}
    \mathbf{s}_n = \mathcal{G}_\theta \left( \tau(\mathbf{s}_{n-1}), \mathcal{C}_n \right),
    \label{eq:sar_generation}
\end{equation}
\noindent where $\mathcal{C}_n = \{c_n^{(1)}, \dots, c_n^{(m)}\}$ denotes the collection of heterogeneous conditions (e.g., images or video clips) falling within the current window. By explicitly conditioning on $\tau(\mathbf{s}_{n-1})$, the generator maintains rigorous pixel-level continuity across segment boundaries.
This auto-regressive mechanism ensures that the distribution of the current segment is strictly aligned with the boundary context of the previous one, maintaining temporal consistency across the entire sequence.

The process above is repeated until all segments are generated. 
Finally, we fuse the overlapping latent frames between adjacent segments and obtain the final latent sequence. 
This sequence is then processed by a VAE decoder, yielding a naturally coherent long video with smooth visual continuity.
It is noteworthy that our segment-wise generation is entirely performed in the latent space, yielding smoother transitions than pixel-based approaches. Moreover, benefiting from the model’s training during the Adaptive Tuning and the Visual Expression SFT, the model inherently maintains visual consistency when extending videos based on conditioning frames, avoiding artifacts such as frame flickering and abrupt jumps.

\section{Experiment}
\label{sec:exp}
In this section, we conduct a systematic evaluation of DreaMontage. 
First, we demonstrate our unique function of arbitrary frame-guided one-shot generation. 
As this capability is largely absent in existing models, we focus on qualitative demonstrations to highlight the narrative coherence and temporal control achieved by our methods.
Second, by formulating first-last and multi-keyframe conditioning as special sub-cases, we benchmark against current SOTA models to evidence our superiority in visual and motion quality through rigorous quantitative and subjective comparisons. 
Finally, we conduct ablation studies to verify the critical contributions of optimizations in our work.

\subsection{Experimental Settings}
\noindent\textbf{Implementation Details.}
As demonstrated in previous sections, our training pipeline comprises three progressive stages.
In the Adaptive Training stage, the filtered one-shot dataset contains 300k filtered video clips, on which the model is trained for 30k steps.
In the Visual Expression SFT stage, we collect nearly 1k high-quality samples and the model is trained for another 15k steps.
Finally, in the Tailored DPO stage, we construct 1k preference pairs for each task and train the model for 10k steps. 
For the super-resolution, we train two specialized models (based on the same DiT architecture) targeting 720p and 1080p resolutions, respectively.
% For all stages, we employ the AdamW optimizer with a constant learning rate of $1 \times 10^{-6}$. 
% All trainings are conducted on 64 Nvidia-A100 GPUs with a global batch size of 128.

\noindent\textbf{Evaluation Dataset.} 
To comprehensively assess the generation capabilities across varying complexities, we curate a large-scale internal test set covering a wide spectrum of common scenarios, encompassing diverse themes, styles, subjects, and semantic contents. 
The target video durations in this dataset range from 5 to 60 seconds, where each test sample is originally constructed with complex sequences of multiple image and video conditions. 
Furthermore, to make fair comparisons with existing models that lack support for arbitrary-frame conditioninig, we derive two specialized subsets from this foundational dataset: 
a \textit{Multi-Keyframe Benchmark}, which excludes video insertion to focus on multi-keyframe conditioning, 
and a \textit{First-Last Frame Benchmark}, which retains only the initial and final frames to align with the standard first-last conditioning setting.

\noindent\textbf{Evaluation Metrics.}
To comprehensively evaluate perceptual quality and generation capabilities, we employ the Good/Same/Bad (GSB) protocol for human evaluation. 
In this pairwise comparison setting, human experts are presented with two videos side-by-side generated by the models under comparison, sharing identical inputs. 
To ensure fairness, the display order is randomized and blindly annotated. 
Evaluators are instructed to judge the results across four key dimensions: Visual Quality, Motion Effects, Prompt Following, and Overall Preference. 
For each pair and dimension, the evaluator assigns a rating of "Good" if the evaluated model outperforms the counterpart, "Bad" if it falls short, or "Same" if the quality is indistinguishable. 
For each comparison pair, by aggregating all rating data, we obtain the counts of Wins, Losses, and Ties. The final GSB score is then calculated as (Wins - Losses) / (Wins + Losses + Ties).
This method allows us to quantify the preference rate and capture subtle nuances in generation capabilities that automated metrics may overlook.

\subsection{One-shot Video Generation}
\begin{figure}[!ht]
    \centering
    \includegraphics[width=\linewidth]{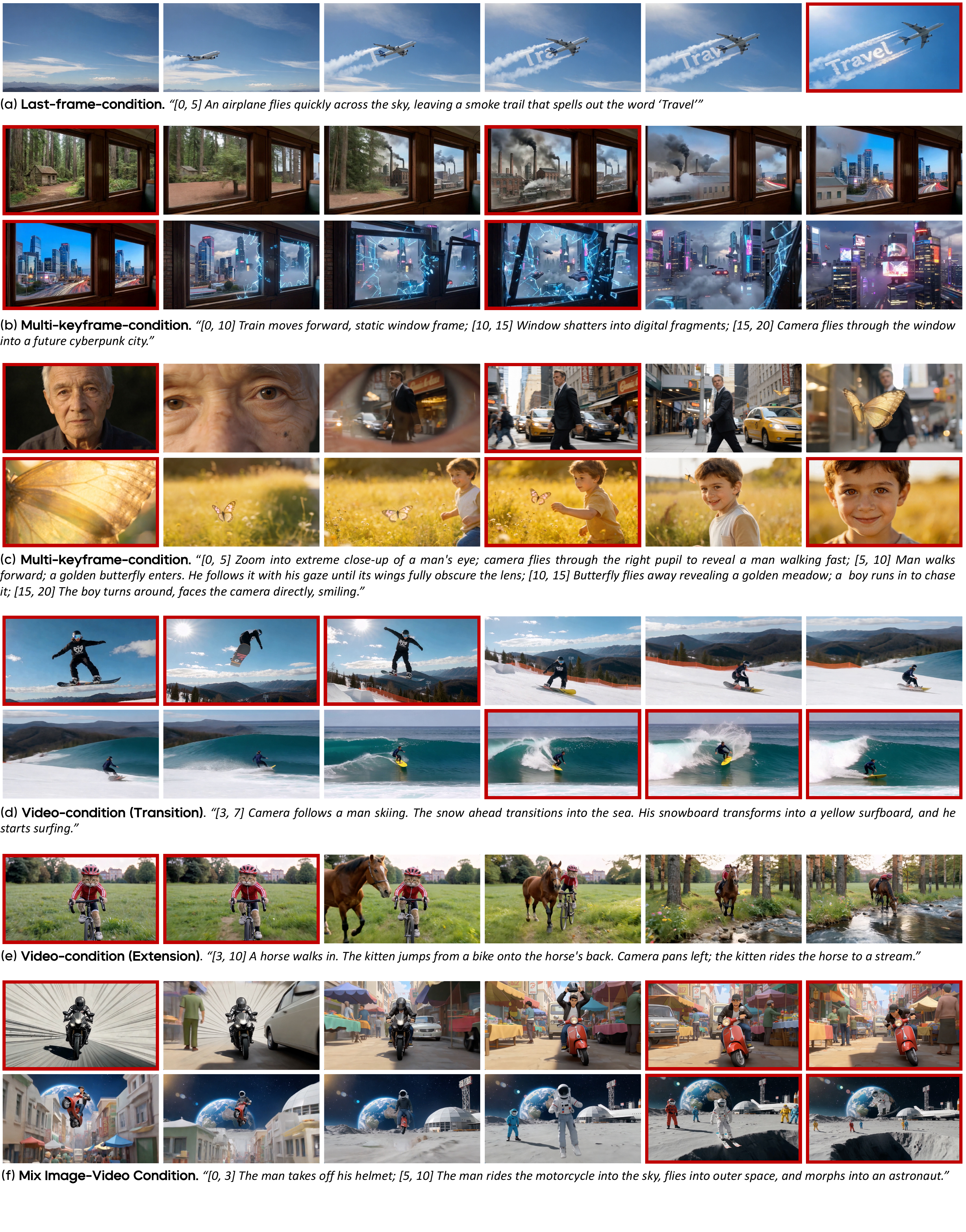}
    \vspace{-30pt}
    \caption{\textbf{Qualitative visualization of arbitrary frame-guided generation.} 
            {\color{red}{Red-bordered}} frames denote the user-provided conditions anchored at specific timestamps, while the other frames represent the generated content.
            }
\label{fig:qualitative_results}
    \label{fig:placeholder}
\end{figure}

We qualitatively evaluate the core capability of DreaMontage: generating long-duration, one-shot videos guided by arbitrary intermediate conditions. 
As visualized in Fig.~\ref{fig:qualitative_results}, we present six distinct scenarios ranging from single-image constraints to complex mixed-media narratives. 
The frames marked with red borders represent user-provided conditions — anchored at precise timestamps.

\noindent\textbf{Complex Narrative Construction.} 
Rows (b) and (c) demonstrate the model's ability to "connect the dots" between disparate visual concepts. 
In the \textit{Multi-keyframe} setting (b), DreaMontage generates a seamless transition from a realistic train interior to a futuristic cyberpunk city, naturally handling the shattering of the window as a bridge between the two distinct styles. 
More strikingly, in row (c), the model executes a sophisticated "match cut" and continuous zoom: transitioning from an extreme close-up of an eye, flying through the pupil to reveal a street scene, and eventually shifting to a meadow. 
This proves that our method can handle drastic scale changes and semantic jumps without producing the morphing artifacts or ghosting effects common in baseline interpolation methods.

\noindent\textbf{Video-Guided Transition and Extension.} 
Beyond static images, DreaMontage excels at handling video clips as conditions. 
Row (d) illustrates a \textit{Video-condition (Transition)} task, where the model takes a skiing clip as one of the conditions and logically evolves the environment from snow to sea, transforming the athlete's motion into surfing (another condition) while preserving physical momentum. 
In the \textit{Extension} case (e), given a starting clip of a cat riding a bike, the model hallucinates a complex interaction where the cat jumps from the bike to the back of a horse, effectively extending the narrative scope while maintaining character consistency.

\noindent\textbf{Flexible Hybrid Control.} 
Finally, row (f) showcases the \textit{Mix Image-Video Condition}, where a static image and two dynamic clips are placed at the start, the middle and the end of the timeline respectively. 
DreaMontage successfully generates a coherent trajectory, depicting the rider taking off a helmet, ascending into the sky, and morphing into an astronaut, thereby translating a static storyboard into a vivid, multi-stage cinematic experience.

Please refer to our \href{https://dreamontage.github.io/DreaMontage}{project page} for more high-resolution video demonstrations. 

\subsection{Comparison with existing Methods}
To the best of our knowledge, no existing method supports inserting video clips as intermediate conditions. 
Instead, current approaches are restricted to multi-keyframe or first-last frame conditioning, both of which constitute special cases of our DreaMontage. 
Therefore, we benchmark against state-of-the-art models under these two respective sub-settings to ensure a fair comparison.

\begin{figure}[t]
    \centering
    \includegraphics[width=\linewidth]{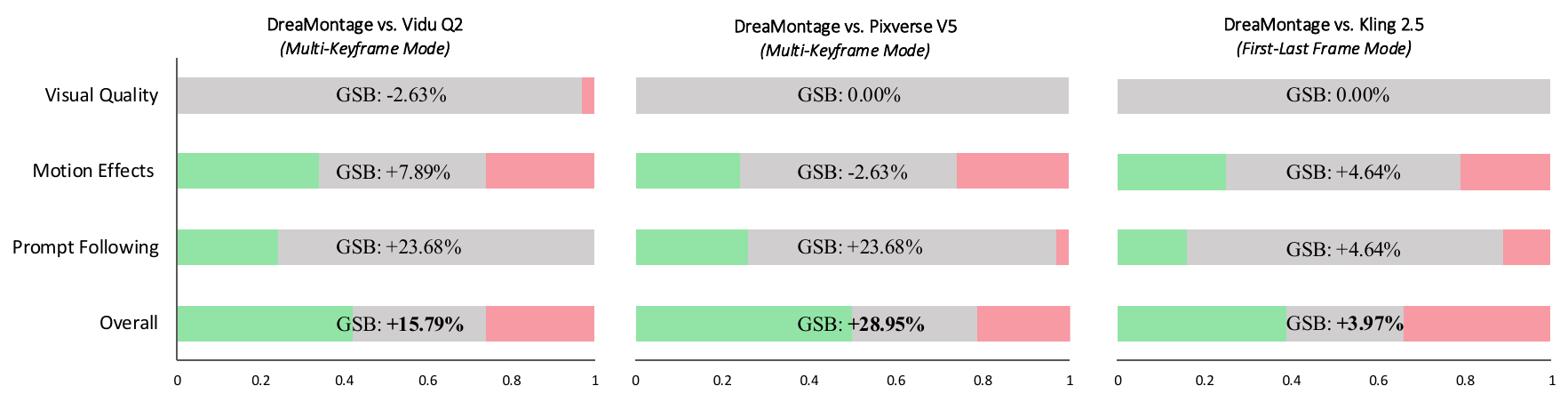}
    \caption{Quantitative comparison with state-of-the-art methods.
    {The \color{green}{\textit{green}}} bars represent the percentage of cases where our result is visually superior, {\color{red}{\textit{red}}} favors the competitor, and {\color{gray}{\textit{gray}}} indicates comparable quality. Reported numbers are the GSB scores.}    
    \label{fig:comparison_sota}
\end{figure}

\noindent\textbf{Multi-Keyframe Mode.}
We benchmark DreaMontage against leading models capable of handling multi-keyframe conditions: Vidu Q2~\cite{shengshu2025vidu} and Pixverse V5~\cite{pixverse2025v5}.
As shown in Fig.~\ref{fig:comparison_sota}, DreaMontage achieves a commanding lead in overall preference, surpassing Vidu Q2 by +15.79\% and Pixverse V5 by +28.95\%.
Notably, the most significant gap is observed in prompt following, where our model outperforms both competitors by a substantial margin of +23.68\%. 
This confirms that our Adaptive Tuning and Visual Expression SFT enable DreaMontage to faithfully respect complex user instructions and semantic constraints, whereas counterparts often struggle to align the generated content with the text prompts in multi-condition settings.
Regarding motion effects, our model demonstrates superior performance compared to Vidu Q2 (+7.89\%) and remains competitive with Pixverse V5 (-2.63\%), striking a robust balance between dynamic movement and stability.
While visual quality shows a slight trade-off compared to Vidu Q2 (-2.63\%) and a tie with Pixverse V5 (0.00\%), the decisive advantage in overall user preference indicates that our model's superior narrative coherence and instruction adherence are more critical for one-shot video generation tasks.

\noindent\textbf{First-Last Frame Mode.}
We compare our approach with the widely recognized model Kling 2.5~\cite{kuaishou2024kling}.
Although Kling 2.5 is a formidable contender renowned for its high-definition generation, our model proves to be robust.
The GSB statistics reveal a draw in visual quality (0.00\%), suggesting that DreaMontage matches the visual fidelity of top-tier commercial models.
However, we secure consistent wins in both motion effects (+4.64\%) and prompt following (also +4.64\%).
Consequently, DreaMontage achieves a higher overall preference (+3.97\%), confirming that it excels not only in its unique multi-frame capability but also maintains SOTA-level performance in standard first-last frame generation task.
% \begin{table}[t]
%     \centering
%     \caption{Quantitative comparison with state-of-the-art methods. \textit{MF} and \textit{FL} denote the \textit{Multi-Keyframe} and \textit{First-Last Frame} conditioning modes, respectively. Numbers represent the GSB score (\%).}
%     \label{tab:comparison_sota}
%     \footnotesize 
%     \setlength{\tabcolsep}{3.5pt}
    
%     \begin{tabular}{lccccc}
%         \toprule
%         Comparison Setting & Visual Quality & Motion Effects & Prompt Following & Overall \\
%         \midrule
%         % Row 1: Vidu Q2
%         Ours \textit{vs.} Vidu Q2 (\textit{MF}) & -2.63 & +7.89 & +23.68 & \textbf{+15.79} \\
%         % Row 2: Pixverse V5
%         Ours \textit{vs.} Pixverse V5 (\textit{MF}) & 0.00 & -2.63 & +23.68 & \textbf{+28.95} \\
%         \midrule
%         % Row 3: Kling 2.5 (使用 midrule 分隔不同模式，逻辑更清晰)
%         Ours \textit{vs.} Kling 2.5 (\textit{FL}) & 0.00 & +4.64 & +4.64 & \textbf{+3.97} \\
%         \bottomrule
%     \end{tabular}
% \end{table}

\subsection{Ablation Study}
\noindent\textbf{Effectiveness of Visual Expression SFT.}
As presented in Table~\ref{tab:ablation}, comparing the SFT-enhanced model against the Base model reveals that the primary contribution of this stage lies in motion dynamics. %rather than static image quality.
Specifically, while visual quality remains comparable (0.00\%), the model achieves a remarkable improvement in motion effects (+24.58\%).
This indicates that the SFT stage effectively activates the model's ability to generate vivid, high-magnitude movements.
Coupled with a modest gain in prompt following (+5.93\%), the Visual Expression SFT leads to a significant boost in overall preference (+20.34\%), confirming its necessity for transforming basic generation capabilities into cinematic storytelling.

\noindent\textbf{Effectiveness of Tailored DPO.}
To rigorously assess the impact of our Tailored DPO, we train a unified model using preference pairs from both the "Abrupt Cuts" and "Subject Motion" pipelines. 
The performance gains are evaluated separately on two specialized dimensions to isolate the improvements for each objective.

\noindent\textit{Performance on Abrupt Cuts.}
As shown in the middle section of Table~\ref{tab:ablation}, compared to the SFT baseline, the DPO-enhanced model achieves a GSB score of +12.59\%.
Qualitative analysis confirms that by penalizing negative samples with hard cuts, the model successfully learns to synthesize smoother narrative bridges between disparate conditioning frames, effectively resolving the transition discontinuity issue.

\noindent\textit{Performance on Subject Motion.}
On this specific task, the same DPO model yields a preference score of +13.44\%.
Visual inspection demonstrates that the model effectively generalizes the preference for natural motion, significantly mitigating anatomical distortions and unnatural movements (e.g., sliding feet or impossible limb rotations) that were occasionally observed in the SFT stage.

\noindent\textbf{Shared-RoPE in Super-Resolution.}
The proposed Shared-RoPE mechanism achieves a dominant preference rate (+53.55\%).
As shown in the last line of Table~\ref{tab:ablation}, in the baseline setting (SR Base), the SR model frequently interprets the minor discrepancies between the low-res latent and the high-res condition as semantic conflicts, resulting in severe temporal flickering and color shifts.
By enforcing alignment through shared rotary position embeddings, our method ensures that the high-resolution generation remains strictly aligned with the conditioning signals, thereby eliminating artifacts and ensuring temporal stability in the final output.
\begin{table}[t]
    \centering
    \caption{Ablation study investigating the performance gains attributed to each optimization strategy. Numbers represent the GSB score (\%).}
    \label{tab:ablation}
    \footnotesize 
    \setlength{\tabcolsep}{3.5pt}
    \begin{tabular}{llcccc} 
        \toprule
        \multicolumn{2}{l}{Comparison Setting} & Visual Quality & Motion Effects & Prompt Following & Overall \\
        \midrule        
        \multicolumn{2}{l}{SFT \textit{vs.} Base (only adaptive training)} & 0.00 & +24.58 & +5.93 & +20.34 \\
        \midrule       
        \multirow{2}{*}{SFT+DPO \textit{vs.} SFT} & \textit{Abrupt Cuts} & - & +12.59 & - & +12.59 \\
                                                  & \textit{Subject Motion} & +13.44 & - & - & +13.44 \\
        \midrule        
        \multicolumn{2}{l}{Shared-RoPE \textit{vs.} SR Base} & +53.55 & - & - & +53.55 \\
        \bottomrule
    \end{tabular}
\end{table}

\subsection{Exploration on Applications}
Beyond standard benchmarking, DreaMontage demonstrates immense potential in real-world creative workflows, particularly where precise control over narrative flow and visual consistency is paramount. We highlight three promising application scenarios enabled by our unique multi-condition architecture.

\noindent\textbf{Cinematic Trailer and Montage Creation.}
Traditional video generation models are limited to extending a single image or text prompt, often losing coherence over time. DreaMontage, however, acts as a "neural editor." By accepting a sequence of mixed inputs—such as static character concept art, keyframe illustrations, and pre-existing video clips—it can synthesize a cohesive cinematic trailer.
Users can define the narrative structure by placing specific "anchor" frames or clips at different timelines, and our model seamlessly fills the gaps with semantically aligned transitions. This capability significantly streamlines the pre-visualization (pre-viz) process for filmmakers, allowing them to iterate on storyboards with motion and temporal dynamics instantly.

\noindent\textbf{Infinite Long-Video Generation.}
While current models struggle with degradation in long-form generation, DreaMontage's architecture naturally supports autoregressive video extension without quality decay.
By treating the tail latents of a generated segment as the initial condition for the next, our model can produce theoretically infinite videos while maintaining strict character and stylistic consistency,
% Unlike standard image-to-video models that drift from the original subject, our SFT and DPO strategies ensure that the "identity" of the subject remains robust across multiple generation cycles
making it ideal for creating continuous vlogs, nature documentaries, or endless loop animations.

\noindent\textbf{Game Cutscenes and Dynamic Advertising.}
In the gaming and advertising industries, assets often exist in mixed formats—high-resolution promotional posters (images) and gameplay recordings (videos). DreaMontage can bridge these distinct modalities.
For instance, it can animate a static product poster into a dynamic sequence that seamlessly transitions into actual usage footage, all guided by a unified text prompt.
This "hybrid-condition" capability allows creators to repurpose existing static assets into engaging video content, reducing production costs while ensuring that the generated motion remains faithful to the brand's visual style.
\section{Conclusion}
In this work, we present DreaMontage, a unified framework that democratizes the creation of cinematic one-shot videos. By empowering DiT architectures with arbitrary frame-guided conditioning and a memory-efficient Segment-wise Auto-Regressive strategy, we effectively overcome the rigidity of existing first-last frame paradigms. Our progressive training pipeline — integrating Adapive Tuning, Visual Expression SFT and Tailored DPO — further ensures both narrative coherence and physical plausibility. Extensive evaluations demonstrate that DreaMontage achieves superior controllability and visual fidelity compared to state-of-the-art baselines, providing a scalable and efficient solution for next-generation visual storytelling.
We hope this work serves as a foundational step toward more flexible and expressive storytelling in the era of generative AI.
% \newpage

\clearpage

\bibliographystyle{plainnat}
\bibliography{main}

\clearpage

% \beginappendix
% \input{sections/Appendix}

\end{CJK*}
\end{document}